\documentclass[lettersize,journal]{IEEEtran}
\usepackage{amsmath,amsfonts}
\usepackage{algorithmic}
\usepackage{algorithm}
\usepackage{array}
\usepackage[caption=false,font=normalsize,labelfont=sf,textfont=sf]{subfig}
\usepackage{textcomp}
\usepackage{stfloats}
\usepackage{multirow}
\usepackage{url}
\usepackage{verbatim}
\usepackage{graphicx}
\usepackage{cite}
\usepackage{color}
\usepackage{bbm}
\usepackage{caption}
\usepackage[normalem]{ulem}
\useunder{\uline}{\ul}{}
\def\BibTeX{{\rm B\kern-.05em{\sc i\kern-.025em b}\kern-.08em
    T\kern-.1667em\lower.7ex\hbox{E}\kern-.125emX}}
\usepackage{balance}

\begin{document}

\title{TL-GAN: Improving Traffic Light Recognition via Data Synthesis for Autonomous Driving}

\author{Danfeng Wang \quad Xin Ma \quad Xiaodong Yang
\thanks{Danfeng Wang and Xin Ma are with the College of Information Science and Engineering, Shandong University, China. E-mail: \texttt{wangdanfeng@mail.sdu.edu.cn} and \texttt{max@sdu.edu.cn}}
\thanks{Xiaodong Yang is with QCraft, San Jose, CA 95112 USA. E-mail: \texttt{xiaodong@qcraft.ai}}
\thanks{Corresponding authors: Xin Ma and Xiaodong Yang}}


\maketitle

\begin{abstract}
Traffic light recognition, as a critical component of the perception module of self-driving vehicles, plays a vital role in the intelligent transportation systems. The prevalent deep learning based traffic light recognition methods heavily hinge on the large quantity and rich diversity of training data. However, it is quite challenging to collect data in various rare scenarios such as flashing, blackout or extreme weather, thus resulting in the imbalanced distribution of training data and consequently the degraded performance in recognizing rare classes. In this paper, we seek to improve traffic light recognition by leveraging data synthesis. Inspired by the generative adversarial networks (GANs), we propose a novel traffic light generation approach TL-GAN to synthesize the data of rare classes to improve traffic light recognition for autonomous driving. 
TL-GAN disentangles traffic light sequence generation into image synthesis and sequence assembling. In the image synthesis stage, our approach enables conditional generation to allow full control of the color of the generated traffic light images. In the sequence assembling stage, we design the style mixing and adaptive template to synthesize realistic and diverse traffic light sequences.
Extensive experiments show that the proposed TL-GAN renders remarkable improvement over the baseline without using the generated data, leading to the state-of-the-art performance in comparison with the competing algorithms that are used for general image synthesis and data imbalance tackling. 
\end{abstract}


\section{Introduction}
\IEEEPARstart{I}{n} self-driving systems~\cite{autonomous_driving}, traffic light recognition (TLR) is an essential part of the perception module~\cite{pillar, tracking}. It provides the basic but vital information for the downstream modules including prediction~\cite{prediction_biswas2021,prediction_wegener2021}, decision making and motion planning~\cite{motion_planning_tacs2018,motion_planning_wang2019}. Therefore it is required to obtain the precise traffic light information in real-time, which is closely related to the safety and comfort of autonomous driving. 

\begin{figure}[htbp]
	\centering
    \includegraphics[width=1.0\linewidth]{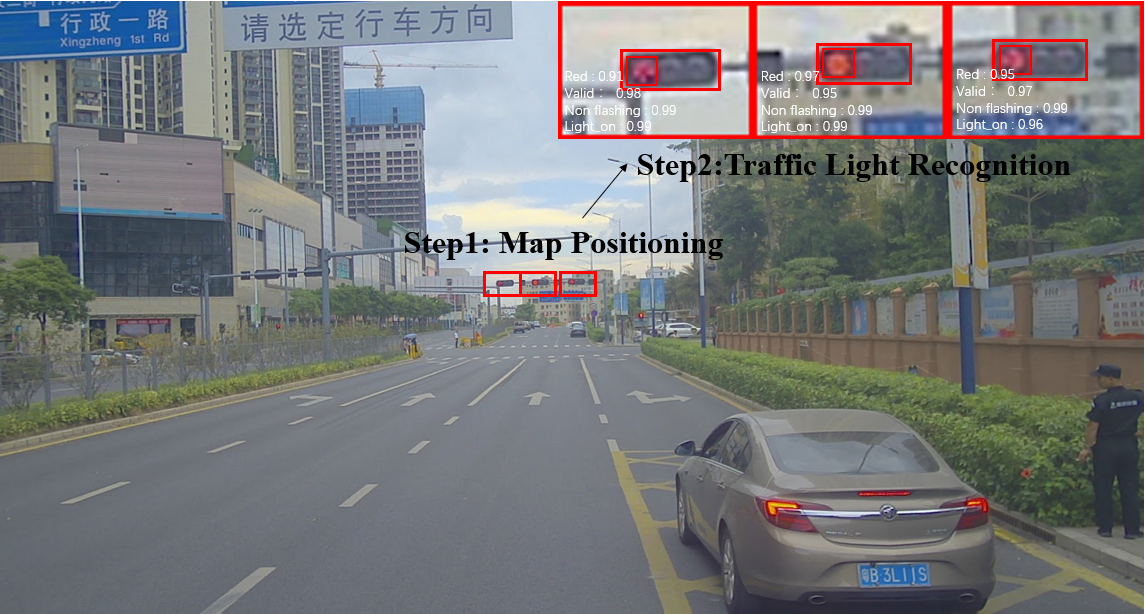}
	\caption{A typical traffic light recognition pipeline of a self-driving vehicle. In the first step, the regions of interest (ROI) can be obtained using the positioning information of traffic lights provided by a high-definition (HD) map. In the second step, the image patches of ROI are input to a vision based TLR model to infer the traffic light status. }
	\label{fig_0}
\end{figure}

Generally, the level 4 self-driving vehicles are equipped with high-definition (HD) maps~\cite{HD_map}, which help to provide the approximate localization of traffic lights on camera images mounted, as shown in Fig.~\ref{fig_0}. First, through the positioning information of traffic lights embedded in the HD map, the regions of interest (ROI) are cropped according to the conversion between the 3D and 2D coordinate systems. Second, a vision based TLR model is used to identify the states of traffic lights in the ROI that are associated with different traffic lanes. 

Nowadays, the deep learning based methods are undoubtedly the mainstream of visual based TLR. 
These methods require massive amount and large diversity of training data to ensure desirable performance.
However, in real-world scenarios, there are a few cases that are not easily available due to the changes of the traffic lights, weather, illumination, background, etc. Thus, the training data naturally exhibits an imbalanced distribution~\cite{long-tailed}, resulting in deteriorated performance on rare classes. To mitigate this problem, there are traditionally two groups of methods that are based on data sampling~\cite{re-sampling} and loss adjusting~\cite{re-weighting}. In order to compensate the rare classes, these methods make use of the imbalanced data distribution to over-sampling or re-weighting to strengthen the learning for rare classes. Although such methods are relatively easy to realize, they seldom capture the semantically meaningful variations in the real data distribution. 

For instance, traffic lights are active most of the time in real scenarios and occasionally off or flashing, resulting in far more non-flashing traffic lights in the collected data than flashing ones. A flashing traffic light sequence is typically composed of alternate light-off (\texttt{inactive}) state and light-on (\texttt{red}/\texttt{yellow}/\texttt{green}) state at different frequencies. 
\begin{figure*}[htbp]
	\centering
    \subfloat[Real]{
    \includegraphics[width=0.5\linewidth]{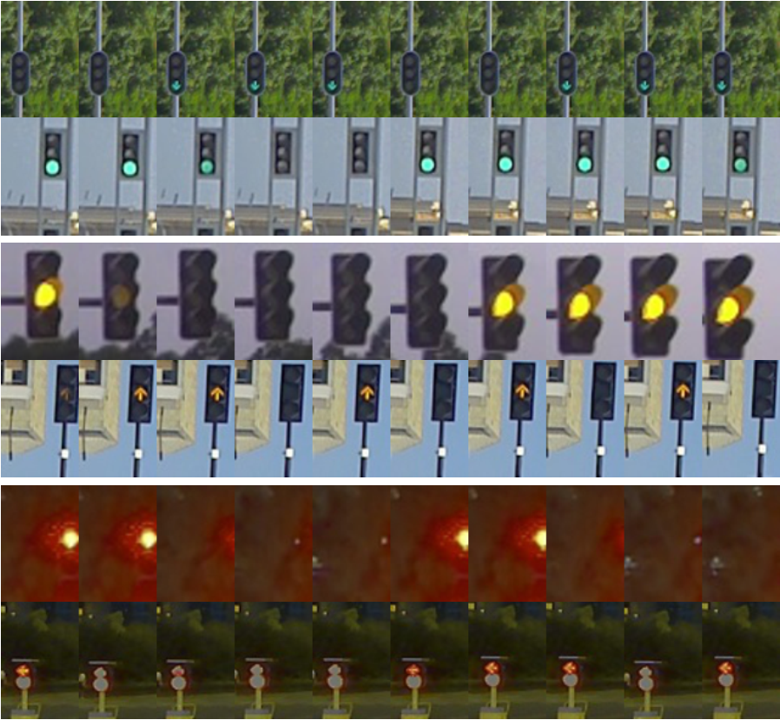}
    }
    \subfloat[Synthesized by TL-GAN]{
    \includegraphics[width=0.5\linewidth]{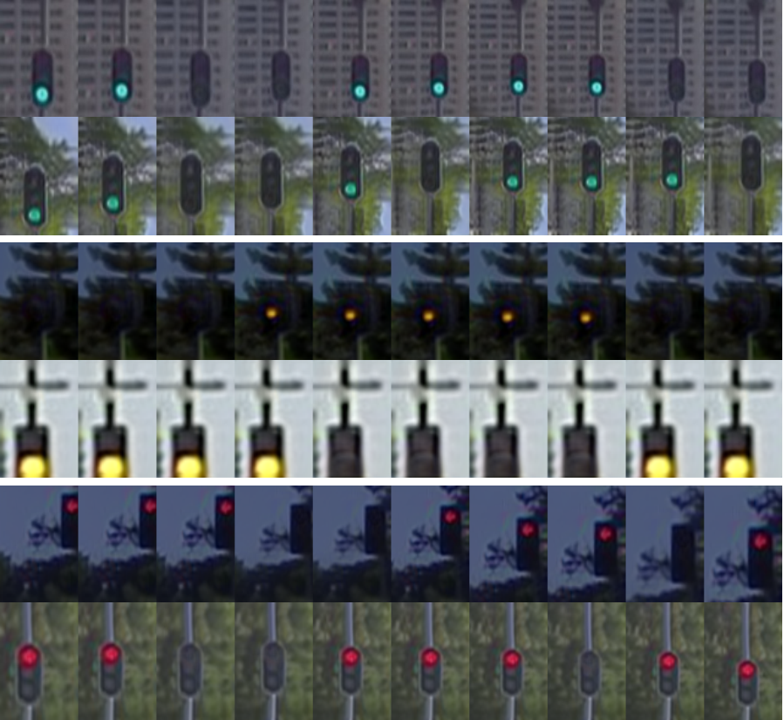}
    }
	\caption{Examples of flashing traffic light sequences. (a) demonstrates the flashing sequences collected in the real data. (b) shows the synthesized flashing sequences generated by our approach.}
	\label{fig_1}
\end{figure*}

As shown in Fig.~\ref{fig_1}, traffic lights in a flashing sequence demonstrate varied color and brightness but similar background. Unfortunately, the conventional methods of data sampling and loss adjusting are incapable of capturing such variations in time dimension. Thanks to the ability of modeling the distribution of real data, the generative adversarial network (GANs)~\cite{gan} have recently attracted attentions in solving the data imbalance problem. 
In this paper, we propose TL-GAN, a new traffic light generation approach, to augment the imbalanced training data by generating synthetic traffic light sequences that can be used to improve the generalizability of recognition models.
\begin{figure}[htbp]
\centering
\includegraphics[width=0.8\linewidth]{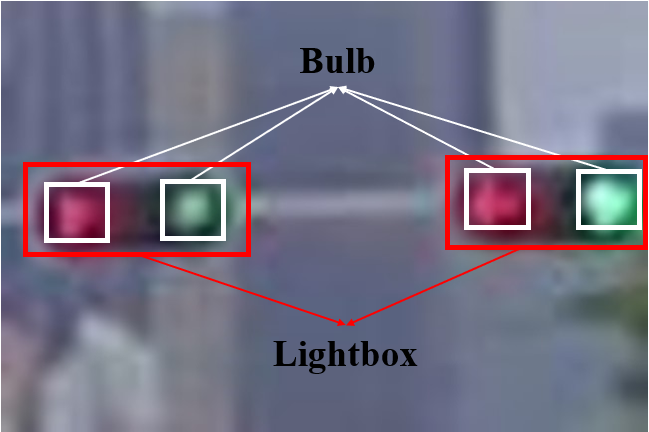}
\centering
\caption{Illustration of the scenario that multiple traffic lights are captured in a single image patch, and the bulb and lightbox regions. When encountering this scenario, it is necessary to detect the multiple lightboxes to confirm which one corresponds to the lane where the self-driving vehicle is located. Besides, when two bulbs in a lightbox are simultaneously active, this means that the traffic light is at a transition state, for example, from green to red. For this case, it is necessary to know the color of each active bulb to understand the traffic light state.}
\label{fig_detection}
\end{figure}
However, it is challenging for GANs to generate realistic and diverse traffic light sequences. On one hand, there are various changes within a flashing sequence: the color and brightness of bulbs can change at different frequencies. On the other hand, there are regions that remain almost constant, such as the background and lightbox. 
It is nontrivial for the image based GANs to achieve the above two aspects in a generated image sequence.
If using the video based GANs, it tends to produce poor quality and low diversity of synthesized video sequences because of the complex flashing patterns and limited flashing samples in real data. 
In light of these observations, we propose a traffic light generation approach called TL-GAN, which disentangles the traffic light sequence generation into two stages: image synthesis and sequence assembling.
Specifically, the first stage performs conditional generation of single traffic light images. We make traffic light color controllable by combining the class embedding with the original latent code. Based on each generated image, the second stage assembles a sequence through manipulating the state of the generated traffic light to simulate various flashing effects. This allows us to only modify the foreground (i.e., the bulb area) while leaving the background unaltered (or slightly changed to imitate the ego-motion of a self-driving vehicle).        

We demonstrate the real and synthesized traffic light data in Fig.~\ref{fig_1}, where the sequences generated by TL-GAN present great realism and diversity.   
Extensive qualitative and quantitative experiments show that TL-GAN outperforms the state-of-the-art generative methods in terms of traffic light synthesis. More importantly, by augmenting the training data with the synthetic data generated by TL-GAN, our TLR model remarkably outperforms the baseline and further delivers better performance than the methods that are specifically designed to tackle the data imbalance problem.     

As for TLR, we develop a traffic light detection model rather than a simple classification model. This is because when adjacent traffic lights are close or there is a certain deviation in the map positioning, multiple traffic light may appear in a single cropped image patch, as shown in Fig.~\ref{fig_detection}. Moreover, a traffic light may have two bulbs active at the same time during state transition. To be more robust to deal with these cases, a TLR model should be able to output the states of all traffic lights and active bulbs. Therefore, our TLR involves bulb and lightbox detection instead of simple image patch classification. 

\begin{figure}[htbp]
	\centering
    \includegraphics[width=0.98\linewidth]{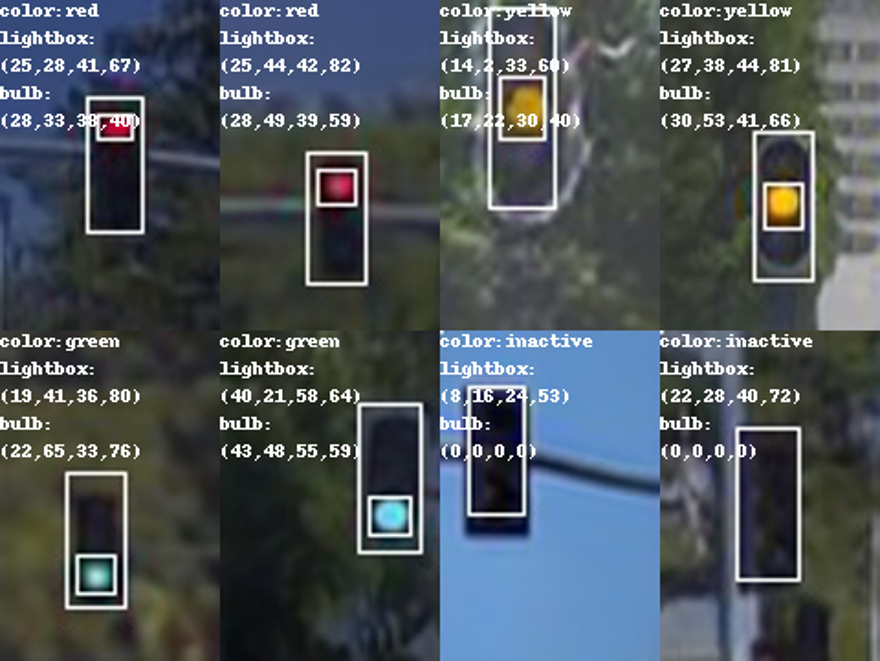}
	\caption{Examples of the generated images with automatic annotations, including the color and bounding boxes of bulb and lightbox.}
	\label{fig_anno}
\end{figure}

Accordingly, TL-GAN can automatically yield rich annotations of the generated data, which can be eventually used to improve the traffic light detection training. As shown in Fig.~\ref{fig_anno}, in addition to the color labels, the bounding boxes of both bulb and lightbox can be produced. These automatic labeling outputs largely facilitate the detection task as aforementioned.
\begin{figure*}[htbp]
\centering
\subfloat[Red]{
\includegraphics[width=0.24\linewidth]{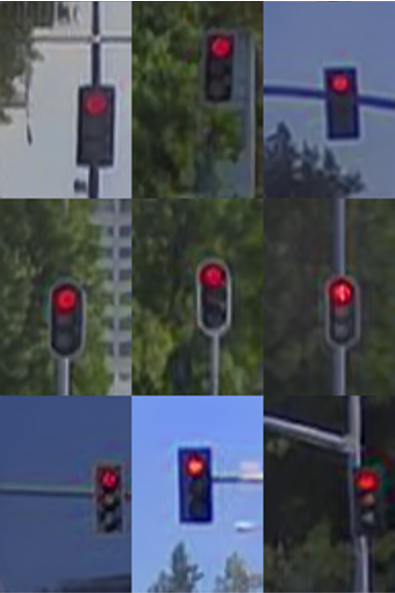}
}
\subfloat[Yellow]{
\includegraphics[width=0.24\linewidth]{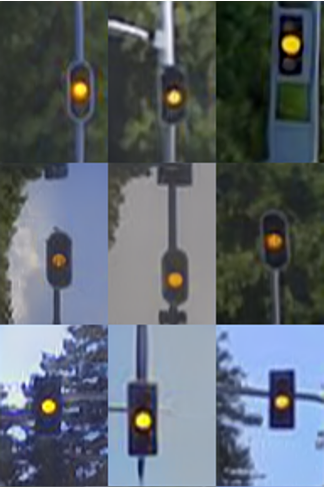}
}
\subfloat[Green]{
\includegraphics[width=0.24\linewidth]{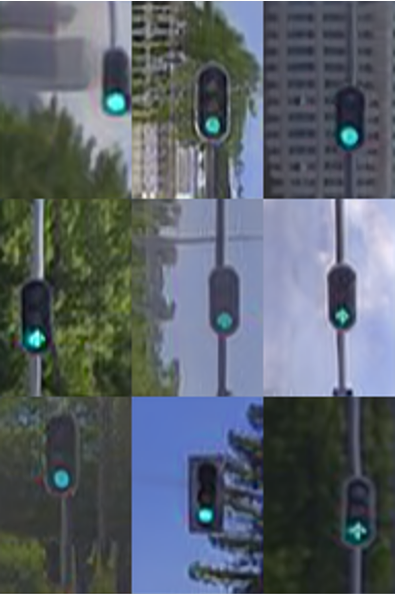}
}%
\subfloat[Inactive]{
\includegraphics[width=0.24\linewidth]{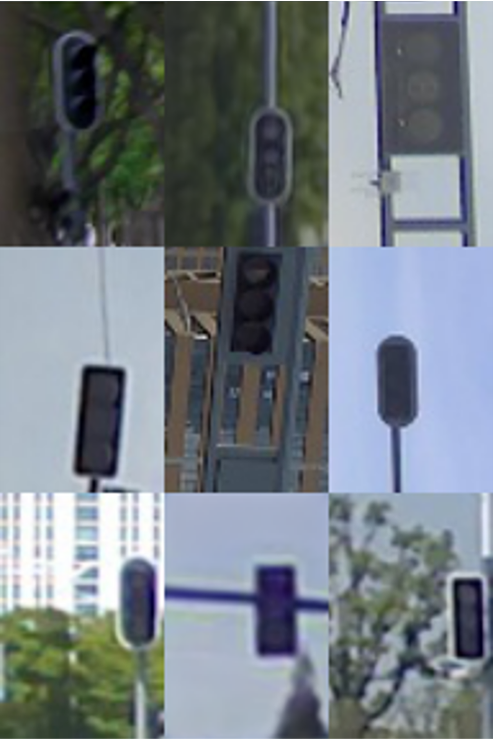}
}%
\centering
\caption{Examples of the generated images by TL-GAN, which can well control the traffic light color. (a-d) show the examples of four classes. The position of the bulb corresponds correctly to the class the traffic light belongs to. Each image is generated with a different latent code.}
\label{fig_2}
\end{figure*}

Our main contributions of this paper can be summarized as follows:
\begin{itemize}
\item{}
We propose a novel two-stage traffic light generation approach TL-GAN, which disentangles traffic light sequence generation into image synthesis and sequence assembling. TL-GAN is shown to be able to generate realistic and diverse samples which are difficult to collect in real scenarios. 
\item{} 
We create a new traffic light dataset including 1440K synthetic sequences, each of which contains automatically labeled class and bounding box information. This dataset will be released to facilitate the research of TLR for real-world autonomous driving.
\item{}
We develop an effective and efficient TLR model by leveraging on the generated data. Our model achieves state-of-the-art performance compared to various competing algorithms, while the inference latency is only 3.8ms running on an NVIDIA 3080 GPU.  
\end{itemize}

The remainder of this paper is organized as follows. Section 2 reviews the related work. Section 3 describes the proposed traffic light generation approach TL-GAN including image synthesis and sequence assembling in detail. Section 4 presents extensive experimental results and analysis. In section 5, we summarize this work and elaborate future directions.

\section{Related Work}
\subsection{Image and Video Generation}
Numerous image and video generation works~\cite{gan, stylegan2019, BigGAN, WGAN-GP, mocogan, dancing} have been proposed and applied in a wide range of vision tasks, such as style transfer~\cite{gan_styletrandfer_1, gan_styletrandfer_2}, super resolution~\cite{Super-Resolution_1, Super-Resolution_2}, object re-identification~\cite{dgnet, dgnetpp}, etc. Meanwhile, GANs provide a feasible way to augment data in few-shot learning tasks~\cite{fewshot}. As for autonomous driving, most methods focus on street view synthesis~\cite{street_1, street_2}, traffic sign generation~\cite{trafficsign_1, trafficsign_2} and style transfer~\cite{styletrasfer_1, styletrasfer_2}. Few works~\cite{trafficlight_gan_2, trafficlight_gan_1} are designed for traffic lights. Mukherjee et al.~\cite{trafficlight_gan_2} proposed a generative method to change the background of traffic light images from day to night. However, simply changing background is apparently insufficient to capture the complex variations of traffic lights. 
Hassan et al.~\cite{trafficlight_gan_1} built a GAN model to obtain the semantics of traffic lights and their surroundings in each image and leverage this model to augment the training data. However, they also mentioned that their augmented images may not be realistic-looking. 

In this paper, we adopt GANs to synthesize photo-realistic traffic light sequences for improving TLR. Our proposed TL-GAN is shown to be able to generate diverse traffic light data with automatically labeled class and bounding box information, which resolves the data imbalance of the rare classes such as yellow, inactive and flashing traffic lights.

\subsection{Data Imbalance Handling}

\textbf{Loss Adjusting}. A variety of methods have been developed to assign different losses to different training samples for each class. Adjusting loss can vary at class level for matching a given data distribution and improving the generalization of rare classes. A more fine-grained loss adjusting can be achieved at sample level including focal loss~\cite{focal_loss}, range loss~\cite{range_loss} and class-balance loss~\cite{classbalance_loss}. He et al.~\cite{focal_loss} proposed a focal loss to handle the data imbalance problem by adaptively assigning more weights to hard samples and less weights to easy samples. Lin et al.~\cite{range_loss} designed a range loss to reduce overall intra-personal variations while enlarging inter-personal differences in each mini-batch simultaneously. Cui et al.~\cite{classbalance_loss} introduced a loss adjusting scheme by using the effective number of samples for each class to re-balance the loss. Although these loss adjusting methods can alleviate data imbalance, their hyper-parameters need to be properly set, which is not trivial to achieve for each new task or dataset.

\textbf{Data Sampling.} Appropriate sampling data from different classes is a basic way to achieve more balanced training distribution. Traditional methods in the field include over-sampling of minority classes~\cite{oversample}, under-sampling~\cite{undersample} of majority classes, and class-balanced sampling~\cite{class-Balanced_sampling} based on the number of samples of each class. Recently, Kang et al.~\cite{kang_Decoupling} proposed a method decoupling the learning procedure into representation and classification, systematically exploring how different balancing strategies affect the final performance. They also empirically showed that using both loss adjusting and data sampling can achieve better performance.

In our paper, to mitigate the imbalanced distribution of training data, we propose TL-GAN to generate realistic and diverse traffic light samples to augment the real data and hence improve the recognition performance in rare classes.

\subsection{Traffic Light Recognition}
\textbf{Map based Models.} HD maps contain various annotations (e.g., position and direction) of traffic lights, which can be used to simplify the difficulty and improve the robustness of TLR. Lindner et al.~\cite{Lindner} applied HD maps to recognize traffic lights based on a three-stage system: detection, tracking and state classification. John et al.~\cite{John} combined HD maps and GPS to constrain the regions of interest where traffic lights are expected to appear. Lucas et al.~\cite{Lucas} utilized a traffic light detector with precise localization provided by HD maps and LiDAR point clouds.

\textbf{Map-Free Models.} DeepTLR was proposed in~\cite{deep_tlr_5} for traffic light detection. Behrendt et al.~\cite{Behrendt} modified YOLO~\cite{yolo} to detect traffic light candidate regions. Pon et al.~\cite{pon} simultaneously detected traffic lights and traffic signs with a modified version of Faster R-CNN~\cite{faster_r-cnn}. 
Although the map-free models do not require HD maps, their performances are susceptible to the interferences by many other factors. For example, it is not easy to distinguish traffic lights from other light sources at night or when they are far way from the self-driving vehicle. It is also quite challenging to associate traffic lights with corresponding lanes without HD maps.

In this work, following the map based models, we focus on TLR for the real-world level 4 autonomous driving applications. Our input traffic light images are the cropped image patches obtained with HD map positioning. We develop a traffic light detection model by leveraging on the real data and synthetic data generated by TL-GAN.

\section{Proposed method}
In this section, we present details of the proposed TL-GAN to generate traffic light sequences for each class, especially the rare classes, thereby obtaining a better TLR model with the synthetic data. Our approach can be distilled down to two stages: image synthesis and sequence assembling. At the first stage, we make the color of generated images controllable by achieving conditional generation. At the second stage, we assemble sequences by manipulating the states of traffic lights to mimic different flashing patterns. Combining the above two stages, we develop a generative approach, which synthesize various traffic light sequences with automatic annotations to assist the training of the traffic light detection model.

\subsection{Image Synthesis}
\begin{figure}[htbp]
\centering
\subfloat[Unconditional model]{
\includegraphics[height=2.1in,width=1.5in]{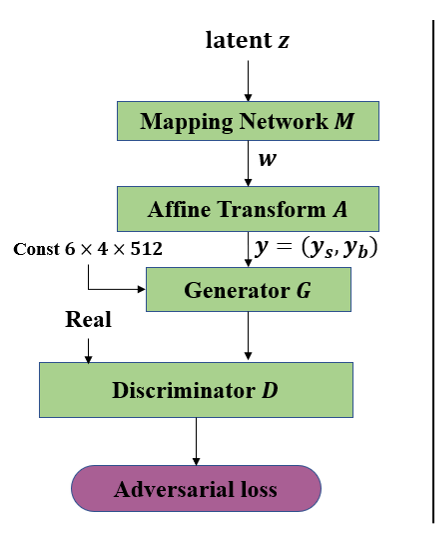}
}
\subfloat[Conditional model]{
\includegraphics[height=2.1in,width=1.5in]{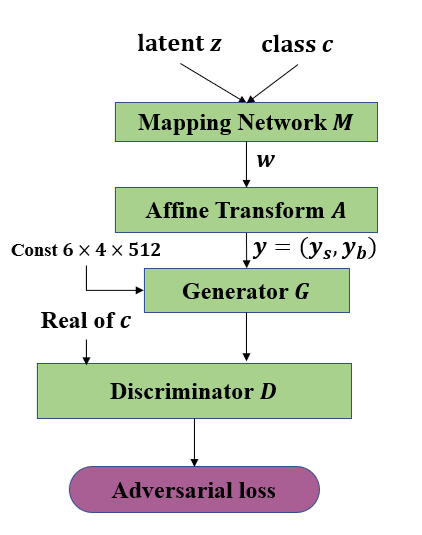}
}
\centering
\caption{Extend (a) an unconditional generative model into (b) a conditional generative model. The latent code $z$ is a random variable sampling from a normal distribution and $c$ is the class identifier. The mapping network $M$ consists of fully connected layers that map the input $z$ or $z$ and $c$ to an intermediate latent space $W$ ($w\in W$). The affine transform module $A$ specializes the style feature $w$ to different style factors $y=(y_s,\ y_b)$ that can be used to control the adaptive instance normalization in the generator $G$. The generated and real images of $c$ are input into the discriminator $D$ to compute the adversarial loss.}
\label{fig_4}
\end{figure}

In the task of TLR, it is essential to recognize the color information of traffic lights correctly. Therefore, for the image synthesis stage, we focus on the precise controlling of the traffic light color of generated images. Our image synthesis model can be built upon any image generative model~\cite{WGAN-GP,BigGAN,PAGAN,FQ-GAN}. In this work, we choose SytleGAN~\cite{stylegan2019} due to its separation of high-level attributes and high generation quality. However, the original StyleGAN is designed for unconditional generation, meaning it is unable to specify or control the traffic light color during the generating process. 
Our strategy towards extending StyleGAN to a conditional generative model is illustrated in Fig.~\ref{fig_4}.

To generate the color-conditional image, for the mapping network $M$, instead of inputting the latent code $z$ directly, we concatenate the latent code $z$ with the embedding of $c$ together: $z^{\prime}=\operatorname{concat}(\operatorname{norm}(z), \operatorname{norm}(\operatorname{embed}(c)))$, where $c\in\{\texttt{r(red)}, \texttt{y(yellow)}, \texttt{g(green)}, \texttt{i(inactive)}\}$ is the color identifier, $\operatorname{norm}(\cdot)$ indicates $\ell_2$ normalization, and $\operatorname{embed}(\cdot)$ transforms the one-hot label vector into a high-dimensional feature space. We then take $z^{\prime}$ as the input of the mapping network $M$ and get the style feature $w=M(z^{\prime})$. $A$ denotes learnable affine transform module, which transforms the style feature $w$ into style factors $y$ to control the adaptive instance normalization (AdaIN)~\cite{AdaIN} operations in the generator $G$. $I_{c}^{s}=G(A(w))$ is the synthesized image, where the subscript $c$ indicates its color identifier and superscript $s$ or $r$ denotes that it is a synthesized or real image. For the discriminator $D$, we compute the output as $D(I, c)=f^T\cdot\operatorname{norm}(\operatorname{embed}(c))$, where $f$ corresponds to the feature vector produced by the last layer of $D$. This operation incorporates the color information into the discriminator and makes the discriminator respect the role of the conditional information in the underlining probabilistic mode. We can then define the adversarial loss $L_{\text{adv}}$ as:
\begin{equation}
L_{\text{adv}}=\mathbb{E}\left[\log D\left(I_{c}^{r}, c\right)+\log \left(1-D\left(G\left(A\left(M(z^{\prime})\right)\right), c \right)\right)\right].
\end{equation}

\begin{figure*}[htbp]
\centering
\subfloat[]{
\includegraphics[height=2.6in,width=1.7in]{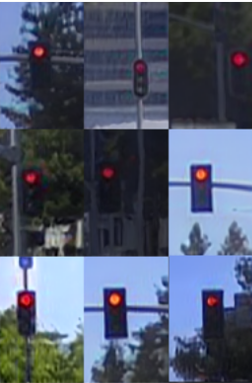}
}
\subfloat[]{
\includegraphics[height=2.6in,width=1.7in]{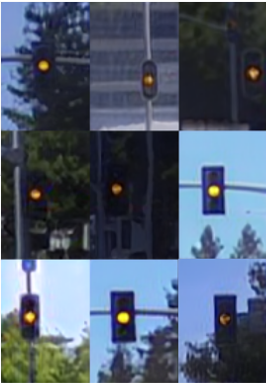}
}
\subfloat[]{
\includegraphics[height=2.6in,width=1.7in]{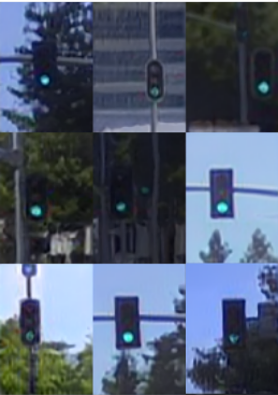}
}%
\subfloat[]{
\includegraphics[height=2.6in,width=1.7in]{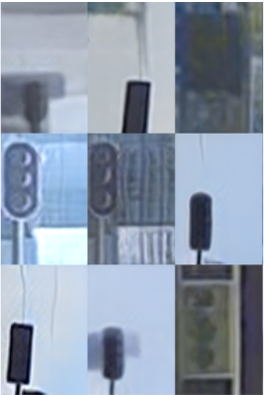}
}%
\centering
\caption{Examples of the active (\texttt{red}/\texttt{yellow}/\texttt{green}) images (a-c) and inactive images (d) generated from the same latent code $z$. Every corresponding images (e.g., the corresponding top left image of each group) are generated from the same latent code with different class embeddings. Note: (1) the generated bulbs of \texttt{red}/\texttt{yellow}/\texttt{green} are accurate: locating from top to bottom, and (2) every corresponding active images have the same background but the inactive one has different background.}
\label{fig_5}
\end{figure*}

With the loss $L_{\text{adv}}$, we are able to make full control of the traffic light color of the generated images, as shown in Fig.~\ref{fig_2}. Moreover, with this design, we can generate such traffic light images that have different colors (\texttt{red}, \texttt{yellow} and \texttt{green}) but share the same background if we input the same latent code with different class embeddings, as demonstrated in Fig.~\ref{fig_5}(a,b,c). However, this is still insufficient to generate a flashing sequence because the backgrounds of inactive traffic light images generated with the same latent code can not be retained, as shown in Fig.~\ref{fig_5}(d). This is due to the fact that the inactive samples of real data have low diversity and inconspicuous color characteristics compared to active classes, resulting in the generation difference.
\subsection{Sequence Assembling}
In order to be able to manipulate the states of traffic lights for flashing sequence generation, we propose style mixing and adaptive template to solve the aforementioned problem that the light-off (\texttt{inactive}) traffic lights and corresponding light-on (\texttt{red}, \texttt{yellow} and \texttt{green}) traffic lights generated from the same latent code have different backgrounds.
The mapping network $M$, affine transform module $A$, generator $G$ and discriminator $D$ are well pre-trained after the aforementioned image synthesis stage. 

\begin{figure*}[htbp]
  \centering
  \includegraphics [width=.9\textwidth] {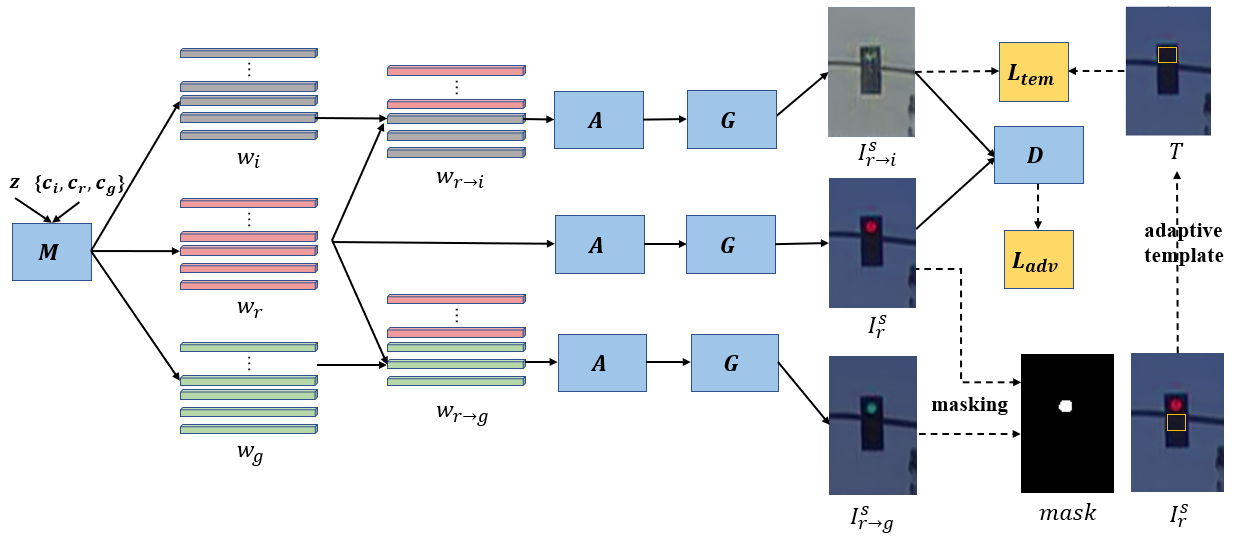}
  \caption{A schematic overview of the sytle mixing and adaptive template design to improve generation of the inactive traffic light image whose background is same as the corresponding active image, which is essential to synthesize a flashing sequence. The mapping network $M$, affine transform module $A$, generator $G$, and discriminator $D$ are well pre-trained in the image synthesis stage. We input the same latent code with different class embeddings into $M$ to produce the style features of $w_{i}$, $w_{r}$ and $w_{g}$. And $w_{r \rightarrow i}$/$w_{r \rightarrow g}$ indicate applying $w_{i}$/$w_{g}$ to the last three affine transform layers of $A$ to produce the last three style factors that relate to traffic light color in the generated image, and $w_{r}$ to the rest affine transform layers of $A$. Using the mixed style factors as the input of $G$, we obtain $I_{r \rightarrow i}^s$ whose quality is low and $I_{r \rightarrow g}^s$ where the position of green bulb is mistaken. To improve the quality of $I_{r \rightarrow i}^s$, we take difference between $I_{r \rightarrow g}^s$ and $I_{r}^{s}$ to have the mask and then obtain the adaptive template $T$ using this mask and $I_{r}^{s}$. Finally, we apply the adaptive template loss to the training to improve the generation of $I_{r \rightarrow i}^s$.} 
  \label{fig_6}
\end{figure*}

As shown in Fig.~\ref{fig_6}, we input the same latent code with different class embeddings, and the mapping network $M$ produces different style features of $w_{i}$, $w_{r}$ and $w_{g}$ corresponding to the \texttt{inactive}, \texttt{red} and \texttt{green} classes. The subscript of $w$ indicates specific class. Among them, the \texttt{inactive} class is required, and the remaining two classes can be randomly selected from \texttt{red}, \texttt{yellow} and \texttt{green}. In this figure, we use the red and green color to illustrate. Each style feature $w$ is input into the affine transform module $A$ to obtain the style factors $\left\{y^{i}=\left(y^{i}_{s}, y^{i}_{b}\right)\right\}$, which control the adaptive instance normalization AdaIN~\cite{AdaIN} in the generator $G$ . 
The resolution of feature maps in $G$ gradually increases from ($6 \times 4$) of the input constant to ($96 \times 64$) of the generated image through upsampling operations, and style factor $y^{i}$ controls the $i$-th feature maps in $G$.
Thus, a larger-indexed style factor controls the higher resolution of the feature maps. Overall, the smaller-indexed style factors control the macroscopic properties of generated images such as shape and background, while the larger-indexed ones control the fine-grained details like color and brightness. Since different style factors control the properties at different levels, in order to only change the traffic light color and keep the background unaltered, we adopt style mixing to switch style factors produced by two different style features and combine them into a new one. 

We empirically find that the last three style factors relate to the control of traffic light color. As illustrated in Fig.~\ref{fig_6}, to change the color of the traffic light from red to inactive or green while keeping other elements unchanged, we use $w_{i}$ and $w_{g}$ as the input to the affine transform module to respectively produce the last three style factors and adopt $w_{r}$ to produce the other style factors. In this paper, we use $w_{r \rightarrow i}$ and $w_{r \rightarrow g}$ to denote this operation and obtain the generated images including $I_{r \rightarrow i}^s$ and $I_{r \rightarrow g}^s$.

\begin{figure}[htbp]
\centering
\subfloat[]{
\includegraphics[height=2.6in,width=1.7in]{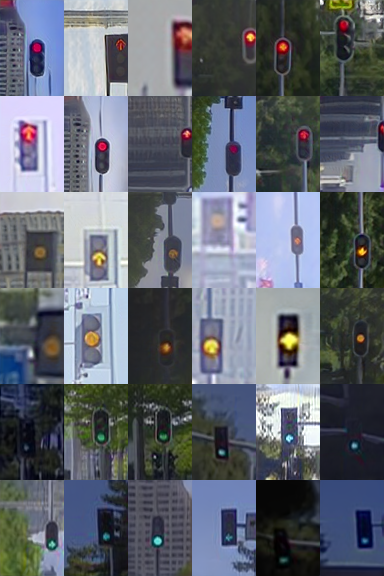}
}
\subfloat[]{
\includegraphics[height=2.6in,width=1.7in]{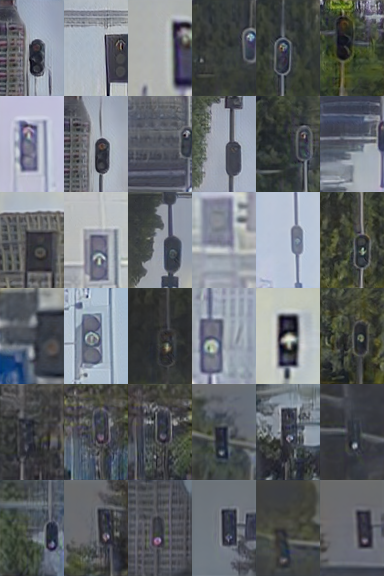}
}
\centering
\caption{Examples of (b) the generated low-quality inactive images which have similar background with (a) the corresponding active ones (e.g., the corresponding top left image of each group).}
\label{fig_7_1}
\end{figure}

Unfortunately, $I_{r \rightarrow i}^s$ performs poorly, not only in the traffic light color but also in the background. More examples are shown in Fig.~\ref{fig_7_1}. We hypothesize that two reasons result in the low quality of the style mixing images of the \texttt{inactive} class: (1) the inactive samples in the training data are limited (3.82\%), and (2) the inactive class does not have definite color information compared with other classes. In addition, although the background of $I_{r \rightarrow g}^s$ is identical to $I_{r}^{s}$ and the traffic light color is successfully changed, the position of the active bulb (\texttt{green}) is mistaken. In reality, the color and position of active bulbs are fixed, i.e., the positions of red, yellow and green bulbs are ordered from top to bottom in a lightbox. It is therefore incorrect for the green bulb to show up at the top position. 

\begin{figure}[hbp]
  \centering
  \includegraphics [width=.4\textwidth] {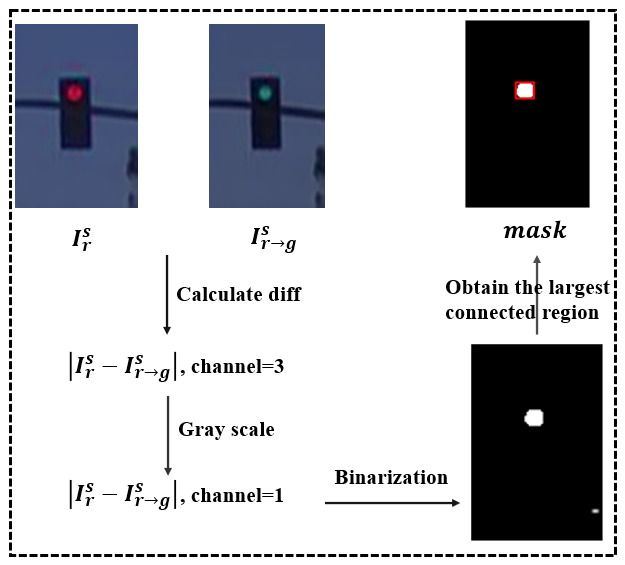}
  \caption{The process of computing the mask for adaptive template. By using style mixing, we obtain $I_{r \rightarrow g}^s$ sharing the same background as $I_{r}^{s}$. Then we calculate the difference between $I_{r \rightarrow g}^s$ and $I_{r}^{s}$. Since the major difference between $I_{r \rightarrow g}^s$ and $I_{r}^{s}$ is the bulb area, we calculate the largest untied region and obtain a mask which can be used to mark the bulb area. Meanwhile, the bounding box of bulb in $I_s^r$ can be output.} 
  \label{fig_8}
\end{figure}

Next we make use of this mis-positioning to produce a mask by taking the difference between $I_{r}^{s}$ and $I_{r \rightarrow g}^s$, as shown in Fig.~\ref{fig_8}. In our paper, the active bulb area is marked as foreground and the area outside the active bulb is background. Hence, the mask can well separate the foreground and background areas of $I_{r}^{s}$. 

\begin{figure}[htbp]
\centering
\subfloat[]{
\includegraphics[height=2.6in,width=1.7in]{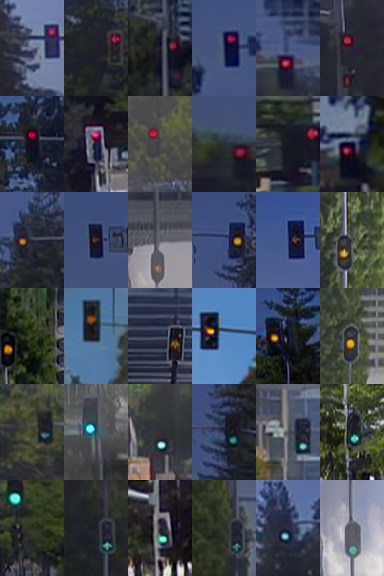}
}
\subfloat[]{
\includegraphics[height=2.6in,width=1.7in]{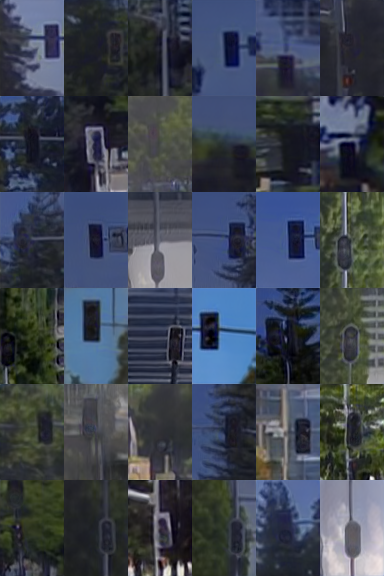}
}
\centering
\caption{Examples of (b) the generated high-quality inactive images which have the same background with (a) the corresponding active ones (e.g., the corresponding top left image of each group).}
\label{fig_10}
\end{figure}

\begin{figure*}[htbp]
\centering
\subfloat[]{
\includegraphics[height=1.75in,width=2.3in]{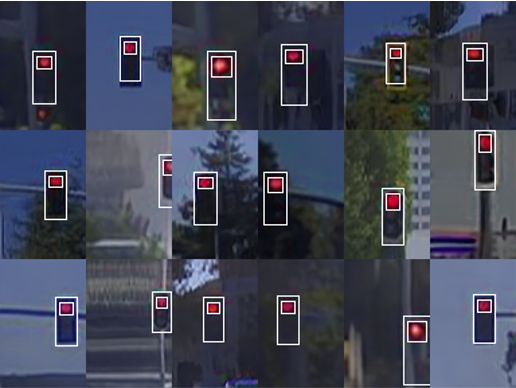}
}
\subfloat[]{
\includegraphics[height=1.75in,width=2.3in]{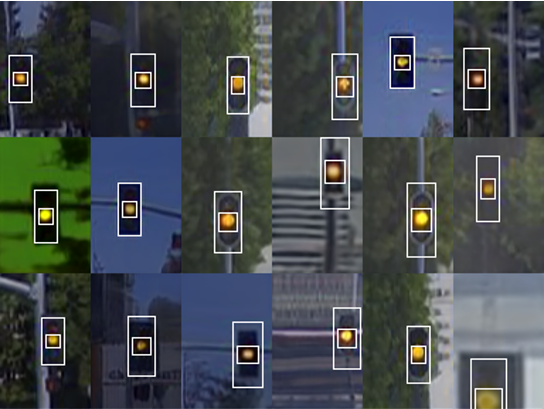}
}
\subfloat[]{
\includegraphics[height=1.75in,width=2.3in]{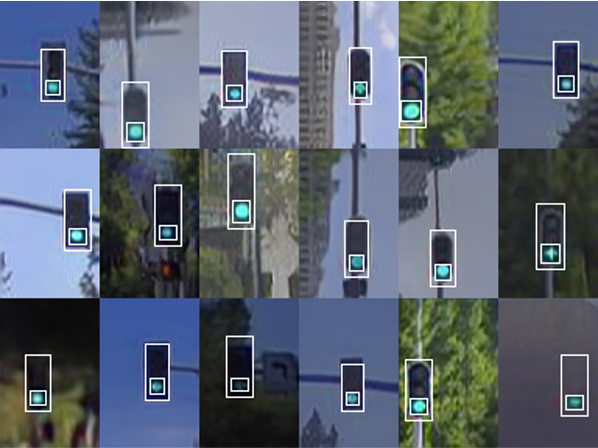}
}%
\centering
\caption{Examples of the generated images with automatically annotated bounding boxes of the bulb and lightbox areas as well as traffic light color information.}
\label{fig_9}
\end{figure*}

To improve the quality of $I_{r \rightarrow i}^s$, we in turn employ the mask to compute an adaptive template $T$ for $I_{r \rightarrow i}^s$ based on the color identifier of $I_{r}^{s}$. As shown in Fig.~\ref{fig_6}, we can easily position the area of active bulb in $I_{r}^{s}$ after acquiring the mask, and then the area of inactive bulb in the lightbox can be inferred according to the position and bulb color. Finally, we obtain an adaptive template $T$ by covering the active bulb area with the light-off area. The template $T$ has the same background as $I_{r}^{s}$, which can be used as the target of $I_{r \rightarrow i}^s$. Thus, we design the adaptive template loss $L_{\text{tem}}$, which uses the pixel-wise $\ell_{1}$ loss to force the generated image $I_{r \rightarrow i}^s$ to be consistent with the adaptive template $T$:
\begin{equation}
L_{\text{tem}}=\mathbb{E}\left[\left\|T-I_s^{r \rightarrow i}\right\|_{1}\right].
\end{equation}
After adding the adaptive template loss, we jointly fine-tune the mapping network, affine transform module, generator and discriminator starting from the first stage model to optimize the overall objective, which is the sum of the following loss terms:
\begin{equation}
L_{\text {total}}\left(M, A, G, D\right)=L_{\text{adv}}+{\lambda}_{\text{tem}}L_{\text{tem}},
\end{equation}
where, ${\lambda}_{\text{tem}}$ is the weight to control the importance of adaptive template loss. With the proposed adaptive template loss, the inactive traffic light images with the similar backgrounds as other classes can be well generated, as shown in Fig.~\ref{fig_10}. Examples of generated flashing sequences are shown in Fig.~\ref{fig_1}(b). 
Meanwhile, the bounding box of the bulb area in a synthetic image can be obtained by calculating the largest connected region as shown in Fig.~\ref{fig_8}. Furthermore, we can proportionally infer the bounding box of the whole lightbox by using its class information and bulb bounding box. Examples of the synthetic images with the automatic generated bounding boxes are shown in Fig.~\ref{fig_9}. 
To simulate the motion of a self-driving vehicle, we randomly crop the generated images (with a large portion to make sure the lightbox is kept after cropping). In addition, to take the different flashing frequency into account, we obtain the statistics of the light-on and light-off frequencies of the real data. Based on this statistical analysis, we alternatively combine the generated active and inactive images that share the same background to synthesize traffic light sequences with various realistic flashing frequencies. Furthermore, with our generation approach, we can fuse the active and inactive images such that the brightness of active traffic lights can be flexibly controlled, i.e., the same-colored bulb can change from bright to dim and then to off, as shown in Fig.~\ref{fig_1}(b).          

\subsection{Traffic Light Recognition}
\begin{figure*}[htbp]
  \centering
  \includegraphics [width=.9\textwidth] {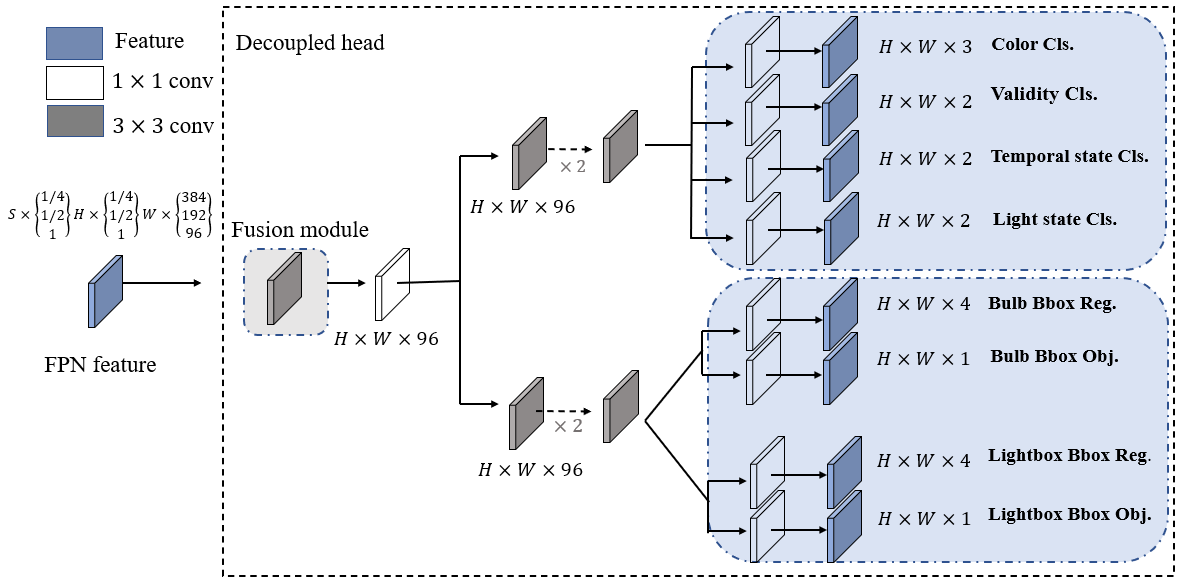}
  \caption{A schematic overview of our traffic light detection model. $S$ is the length of the input sequence, $H=12$ and $W=8$ are the height and the width of the feature map. Feature Pyramid Networks (FPN)~\cite{fpn} feature is the output of backbone network, consiting of proportionally sized feature maps at three levels. For each level of FPN feature, we adopt fusion module to fuse the feature maps of the input sequence on the time dimension. The fusion module first concatenates the feature maps of two consecutive frames in the channel dimension and uses a $3\times3$ conv layer to reduce the number of channels to the original number, and then fuses the fused feature map with the feature map of the next frame. After the fusion module, we adopt a $1\times1$ conv layer to change the the number of fused feature maps' channel to 256. Finally, we add two parallel branches with two $3\times3$ conv layers each for classification and regression tasks respectively.} 
  \label{fig_13}
\end{figure*}

The synthetic data generated by TL-GAN can be used to boost the performance of any object detection model that is adapted for TLR. In this paper, we choose YOLOX-Tiny~\cite{yolox2021} and modify the original model to make it suit for our task.
A schematic overview of our modified model is shown in Fig.~\ref{fig_13}. The shaded parts are our changes including the fusion module (gray) and heads (blue). 

The input to our detection model is a sequence consisting of 10 frames, where the label of last frame is used as the sequence label. Instead of RNNs~\cite{rnns}, we use the fusion module to integrate the features across multiple frames. 
For the classification head, our model predicts a variety of traffic light in detail including validity of lightbox and color, temporal state and light state of bulbs. For the detection head, our model outputs the bounding boxes of bulb and lightbox simultaneously. Our objective function to train the traffic light detection model is defined as: 
\begin{equation}
\begin{aligned}
L &=\lambda_{\text{reg}}^{\text{l}} L_{\text {reg }}^{\text{l}}+L_{\text {obj}}^{\text{l}}+L_{\text {validity}}^{\text{l}} \\
&+L_{\text {obj }}^{\text{b}}+\mathbbm{1}^{\text{active}}[\lambda_{\text{reg}}^{\text{b}} L_{\text {reg}}^{\text{b}}+L_{\text {color}}^{\text{b}}+L_{\text {temporal }}^{\text{\text{b}}}+L_{\text {light}}^{\text{b}}]
\end{aligned}
\end{equation}
where $\lambda_{\text{reg}}^{\text{l}}$ and $\lambda_{\text{reg}}^{\text{b}}$ are weights to control the importance of related loss terms. For lightbox, $L_\text{{reg}}^{\text{l}}$ and $L_{\text{obj}}^{\text{l}}$ are regression loss and objectiveness loss of predicted lightbox. $L_{\text{validity}}^{\text{l}}$ is classification loss of validity which is a attribute of lightbox. 
The lightbox in ground truth can be divided into two types according to its validity, namely \texttt{active} and \texttt{inactive}, depending on whether the bulb in lightbox is on or off.
For the bulb, in addition to calculating regression loss  $L_{\text{reg}}^{\text{b}}$ and objectiveness loss $L_{\text{obj}}^{\text{b}}$, we also need to calculate the classification losses of color $L_{\text{color}}^{\text{b}}$, temporal state $L_{\text{temporal}}^{\text{b}}$ and light state $L_{\text{light}}^{\text{b}}$.
If the validity attribute is \texttt{inactive}, i.e., all bulbs in a lightbox are off more than 1 second, the losses related to predicted bounding box of bulb including $L_{\text{reg}}^{\text{b}}$, $L_{\text{color}}^{\text{b}}$, $L_{\text{temporal}}^{\text{b}}$ and $L_{\text{light}}^{\text{b}}$ do not contribute to the total loss $L$.

Our design of the traffic light detection model can produce the bounding boxes of both bulb and lightbox and predict various traffic light states, including the validity of the lightbox and the color, temporal state and light state of the bulb (see Table~\ref{table_2}).

\section{Experiments}
In this section, we first present the qualitative and quantitative evaluations of the proposed TL-GAN. Then, we apply the generated data to improve the performance of our traffic light detection model. Finally, we conduct extensive comparisons with the methods that are designed to resolve the data imbalance problem.

\subsection{Dataset}
Generative models require a reasonable amount of real images to learn the distribution of the real data. In~\cite{Limited_Data} Karras et al. found that if the amount of real data is very limited, the performance of generative models is poor even with data augmentation. 
In order to enable valid generation and evaluation, we introduce QCraft Traffic Light (QTL), which is a large-scale dataset with annotations of classification, detection and tracking of traffic lights. QTL contains 1.038M sequences, each of which is with a sequence length of 10 frames, and every frame is normalized to a resolution of 75$\times$50. In our experiments, we randomly select 182K sequences as the test set, and the rest is used as the training set.
\begin{table*}[]
\caption{Description of the class definition of the QCraft Traffic Light (QTL) dataset.}
\centering
\scalebox{1.2}{
\begin{tabular}{|c|c|c|c|c|r|}
\hline
Label                        & Color   & Validity    & Temporal   state & Light state  \\ \hline
RED                         & \texttt{red}     & \texttt{active}    & \texttt{non-flashing}     & \texttt{on}                                 \\ \hline
YELLOW                       & \texttt{yellow}  & \texttt{active}    & \texttt{non-flashing}     & \texttt{on}                                 \\ \hline
GREEN                        & \texttt{green}   & \texttt{active}    & \texttt{non-flashing}     & \texttt{on}                                \\ \hline
INACTIVE                     & - & \texttt{inactive} & -    & -                                \\ \hline
RED\_FLASHING\_LIGHT\_ON     & \texttt{red}     & \texttt{active}    & \texttt{flashing}         & \texttt{on}                                    \\ \hline
RED\_FLASHING\_LIGHT\_OFF   & \texttt{red} & \texttt{active} & \texttt{flashing}         & \texttt{off}                                  \\ \hline
YELLOW\_FLASHING\_LIGHT\_ON & \texttt{yellow}  & \texttt{active}    & \texttt{flashing}         & \texttt{on}                                 \\ \hline
YELLOW\_FLASHING\_LIGHT\_OFF & \texttt{yellow} & \texttt{active} & \texttt{flashing}         & \texttt{off}                                \\ \hline
GREEN\_FLASHING\_LIGHT\_ON   & \texttt{green}   & \texttt{active}    & \texttt{flashing}         & \texttt{on}                                 \\ \hline
GREEN\_FLASHING\_LIGHT\_OFF         & \texttt{green} & \texttt{active} & \texttt{flashing}         & \texttt{off}                            \\ \hline
\end{tabular}
}
\label{table_1}
\end{table*}

\begin{table}[]
\caption{Summary of the QCraft Traffic Light (QTL) dataset.}
\centering
\setlength{\tabcolsep}{4mm}{
\begin{tabular}{|c|c|r|r|}
\hline
Attribute                          & Class        & \multicolumn{1}{c|}{Quantity} & \multicolumn{1}{c|}{Ratio(\%)} \\ \hline
                                   & \texttt{red}          & 849450                        & 60.60                          \\ \cline{2-4} 
                                   & \texttt{yellow}       & 43442                         & {3.15}    \\ \cline{2-4} 
\multirow{-3}{*}{Color}            & \texttt{green}        & 485903                        & 35.25                          \\ \hline
                                   & \texttt{active}        & 1349389                       & 96.18                          \\ \cline{2-4} 
\multirow{-2}{*}{Validity}            & \texttt{inactive}     & 53519                         & {3.82}    \\ \hline
                                   & \texttt{non-flashing} & 1306704                       & 94.77                          \\ \cline{2-4} 
\multirow{-2}{*}{Temporal   state} & \texttt{flashing}     & 72091                         & {5.23}    \\ \hline
                                   & \texttt{on}           & 1319983                       & 94.09                          \\ \cline{2-4} 
\multirow{-2}{*}{Light   state}    & \texttt{off}          & 82925                         & {5.91}    \\ \hline
\end{tabular}}
\label{table_2}
\end{table}
More information of this dataset is summarized in Tables~\ref{table_1} and~\ref{table_2}.
Each label has four attributes representing color, validity, temporal state and light state, where \texttt{inactive} denotes that the traffic light is off last more than 1 second, and \texttt{flashing} indicates that the traffic light is in a flashing mode. Each image is annotated with class labels and bounding boxes. Table~\ref{table_2} shows the statistics of color, validity, temporal state and light state. As we can see from the statistics, the data distribution is heavily biased towards \texttt{red} and \texttt{green}, while \texttt{yellow}, \texttt{inactive}, \texttt{flashing} and \texttt{off} account for much fewer proportions of the whole data.

\subsection{Generative Evaluations}
We first qualitatively and quantitatively compare the synthesized results of TL-GAN and other generative methods, including WGAN-GP~\cite{WGAN-GP}, BigGAN~\cite{BigGAN}, PA-GAN~\cite{PAGAN} and FQ-GAN~\cite{FQ-GAN}.

\textbf{Implementation Details.} We implement our approach in PyTorch, and train the models on 2 NVIDIA V100 GPUs.
We set the batch size to 64 and the resolution of generated images to $96 \times 64$. In the image synthesis stage, we train the model for 60K iterations, and set the initial learning rate to 0.0025 and use Adam as the optimizer. In the sequence assembling stage, we fine-tune the model for 40K iterations, and set the initial learning rate to 0.0001. The $\lambda_{\text{tem}}$ is initially set to 0 and and increases linearly to its maximum value of 1 over 2K iterations.

\begin{figure*}[htbp]
\centering
\subfloat[WGAN-GP]{
\includegraphics[height=1.7in,width=1.15in]{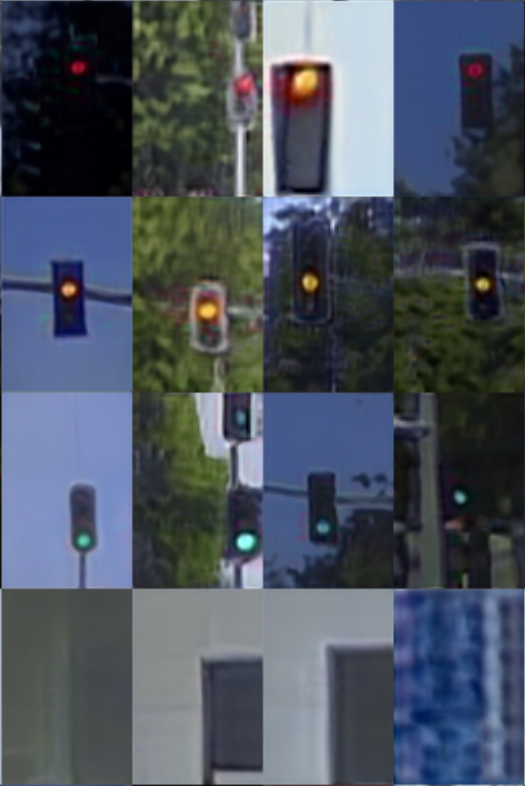}
}
\subfloat[BigGAN]{
\includegraphics[height=1.7in,width=1.15in]{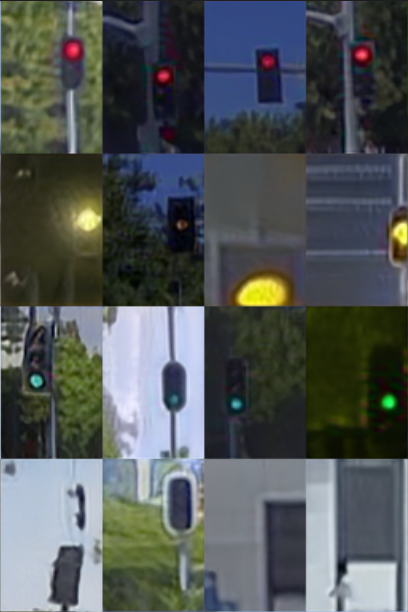}
}
\subfloat[PA-GAN]{
\includegraphics[height=1.7in,width=1.15in]{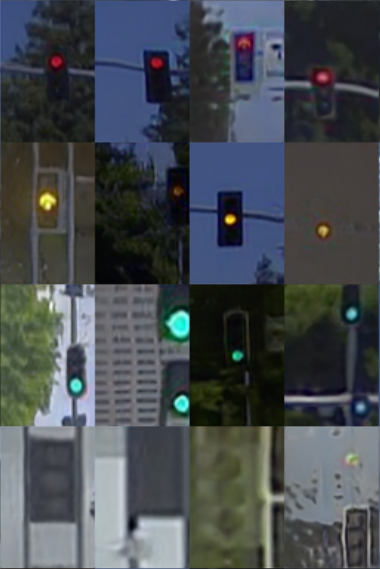}
}
\subfloat[FQ-GAN]{
\includegraphics[height=1.7in,width=1.15in]{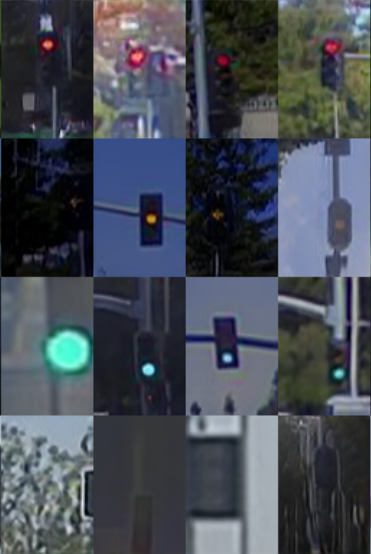}
}
\subfloat[TL-GAN]{
\includegraphics[height=1.7in,width=1.15in]{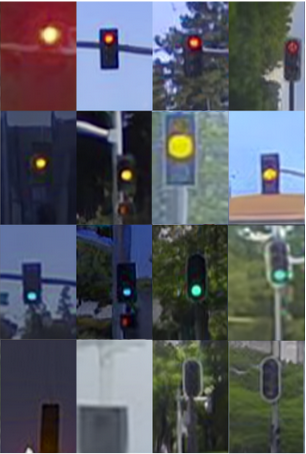}
}
\subfloat[Real]{
\includegraphics[height=1.7in,width=1.15in]{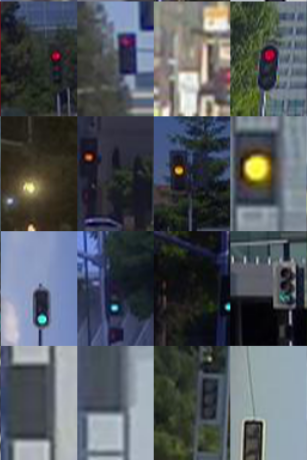}
}
\centering
\caption{Comparison of the real traffic light images and the generated ones using different methods including WGAN-GP~\cite{WGAN-GP}, BigGAN~\cite{BigGAN}, PA-GAN~\cite{PAGAN}, FQ-GAN~\cite{FQ-GAN} and our approach.}
\label{fig_11}
\end{figure*}

\textbf{Qualitative Evaluations}.
As compared in Fig.~\ref{fig_11}, the traffic lights in the images generated by WGAN-GP, BigGAN and PA-GAN have deformed shapes, and the backgrounds contain obvious visual artifacts, which also exist in the images generated by FQ-GAN. In comparison, our generated images are more realistic, and hard to differentiate from the real images in both foreground and background.

\textbf{Quantitative Evaluations}.
Our qualitative observations can be confirmed by the quantitative evaluations. We employ the following two metrics: Fréchet Inception Distance (FID)~\cite{FID} and Kernel Inception Distance (KID)~\cite{KID}.
\begin{itemize}
\item{\bf Fréchet Inception Distance:}
FID is a metric used to measure the realism and diversity of synthesized images by comparing the distribution of generated images with the distribution of real images that are used to train the generative model. 
\item{\bf Kernel Inception Distance:}
KID measures the discrepancy between the distributions of generated and real images using the samples drawn independently from the two distributions. KID is unbiased by design and more descriptive in practice compared to FID.
\end{itemize}

Similar to~\cite{Limited_Data}, we apply the official pre-trained Inception-V3~\cite{inception} to compute FID and KID. As shown in Table~\ref{table_3}, our approach achieves the best results in terms of both FID and KID compared with other generative methods, suggesting the high generation quality of our synthesized images.

\begin{table}[]
\centering
\caption{Comparison of FID and KID to evaluate the generation quality of different generative methods.}
\scalebox{1.0}{
\setlength{\tabcolsep}{6mm}{
\begin{tabular}{|c|c|c|}
\hline
Method  & \multicolumn{1}{c|}{FID$\downarrow$} & KID $\times10^{3}\downarrow$ \\ \hline
Real  & 3.42  & 0.45                     \\ \hline
WGAN-GP~\cite{WGAN-GP} & 33.31                                       & 13.04                     \\ \hline
BigGAN~\cite{BigGAN}  & 17.15                                  & 6.48                      \\ \hline
PA-GAN~\cite{PAGAN}  & 14.90                                    & 5.64                      \\ \hline
FQ-GAN~\cite{FQ-GAN}  & 8.31                                   & 2.17                      \\ \hline
Ours    & \textbf{7.04}                     & \textbf{1.81}             \\ \hline
\end{tabular}}}
\label{table_3}
\end{table}
We make a further comparison for the generation of inactive images sharing with the same background as other active classes, which is the essential step of sequence assembling. In our approach, we utilize style mixing and adaptive template to achieve this goal. As shown in Table~\ref{table_4}, inactive images generated through style mixing only have low quality, like $I_{r \rightarrow i}^s$ illustrated in Fig.~\ref{fig_6} and more examples shown in Fig.~\ref{fig_7_1}(b). According to our analysis above, this is due to the limited samples (small quantity and low diversity) of inactive images collected in the real data. With the adaptive template design, the generation result is significantly improved by enforcing the adaptive template loss in training the same generative model.

\begin{table}[]
\centering
\caption{Evaluation of the generation quality of inactive images.}
\scalebox{1.0}{
\setlength{\tabcolsep}{6mm}{
\begin{tabular}{|l|c|c|c|}
\hline
Method         & FID$\downarrow$  & KID $\times10^{3}\downarrow$ \\ \hline
Real           & 5.05 & 0.83                      \\ \hline
+Style Mixing  & 70.91 & 25.51                      \\ \hline
+Adaptive Template Loss & 8.20 & 2.06                      \\ \hline
\end{tabular}}}
\label{table_4}
\end{table}

\subsection{Improving TLR by Synthesized Data}
As shown in Table~\ref{table_time}, the inference speed of our TLR model is fast, in particular after quantization. As for the detection accuracy shown in Table~\ref{table_5}, the baseline model achieves impressive results on the regular classes, while the results on the rare classes are degraded due to the data imbalance issue.
To improve the performance on the rare classes, our TLR model is first pre-trained on the synthesized data and then fine-tuned using the real data. To reduce the gap between \texttt{active} and \texttt{inactive}, we generate 360K sequences for INACTIVE, 180K sequences for RED, YELLOW and GREEN, and 90K sequences for the rest. 

\textbf{Implementation Details}. We set the sequence length to 10 and batch size to 256. The size of the input image ranges from (72, 48) to (120, 80). The number of training epoch is set as 15. We set the loss weights $\lambda_{\text{reg}}^{\text{b}}=4$ and $\lambda_{\text{reg}}^{\text{l}}=2$ for pre-training on the generated data, and $\lambda_{\text{reg}}^{\text{b}}=3$ and $\lambda_{\text{reg}}^{\text{l}}=5$ for fine-tuning on the real data.

\textbf{Application of Synthesized Data}. During the pre-training on the generated data, our detection model is consistent with the baseline model. We first warm up the learning rate from 0 to the max value 0.04, and then apply the cosine decay scheduler. 
When fine-tuning, to find a suitable initial learning rate, we experiment with different initial learning rates, including 0.02, 0.01 and 0.005.
\begin{table}[]
\centering
\caption{Inference latency of our traffic light detection model with a batch size of 10 on NVIDIA 3080 GPU. }
\scalebox{1.2}{
\setlength{\tabcolsep}{8mm}{
\begin{tabular}{|c|c|}
\hline
Precision & Latency (ms) \\ \hline
FP32 & 10.67 \\ \hline
FP16 & 4.89 \\ \hline
INT8 & 3.88 \\ \hline
\end{tabular}
}}
\label{table_time}
\end{table}

We use $\text{AP}_{50}$, the average precision evaluated at the 0.5 IoU threshold, to evaluate the detection performance. 
Due to the data imbalance problem, we focus on the performance on the rare classes. The results of the fine-tuned model using different initial learning rates are shown in Table~\ref{table_5}. 
We can find that all fine-tuned models using the generated data perform better than the baseline, suggesting that using the data generated by TL-GAN is a valid way to mitigate data imbalance and improve the TLR performance. 

Then, we use different proportions of the synthetic data to pre-train the model and choose 0.01 as the initial learning rate for fine-tuning to explore the impact of using different amounts of the synthetic data.  We use 10\%$\sim$100\% of the generated data for pre-training, and the final evaluation results are shown in Table~\ref{table_6}. Most models with different amounts of synthetic data during pre-training outperform the baseline, further proving the realism and diversity of the synthetic data to resolve the data imbalance problem. 
It can be seen from Table~\ref{table_6} that when the amount of data used exceeds 80\%, the performance tends to be saturated and we hypothesise that this is because the current test set with relatively limited scenarios cannot fully reflect the role of the generated data.

\begin{table*}[]
\centering
\caption{Results of fine-tuning our traffic light detection model (pre-trained on the generated data by TL-GAN) using different initial learning rates.}
\begin{tabular}{|cc|ccccccccc|}
\hline
\multicolumn{2}{|c|}{\multirow{3}{*}{Method}} & \multicolumn{9}{c|}{$AP_{50}$} \\ \cline{3-11} 
\multicolumn{2}{|c|}{} & \multicolumn{3}{c|}{Color} & \multicolumn{2}{c|}{Validity} & \multicolumn{2}{c|}{Temporal state} & \multicolumn{2}{c|}{Light state} \\  \cline{3-11} \multicolumn{2}{|c|}{} & \multicolumn{1}{c|}{\texttt{red}} & \multicolumn{1}{c|}{\texttt{yellow}*} & \multicolumn{1}{c|}{\texttt{green}} & \multicolumn{1}{c|}{\texttt{active}} & \multicolumn{1}{c|}{\texttt{inactive}*} & \multicolumn{1}{c|}{\texttt{flashing}*} & \multicolumn{1}{c|}{\texttt{non-flashing}} & \multicolumn{1}{c|}{\texttt{on}} & \texttt{off}* \\  \hline
\multicolumn{2}{|c|}{Baseline} & \multicolumn{1}{c|}{99.3} & \multicolumn{1}{c|}{98.0} & \multicolumn{1}{c|}{98.9} & \multicolumn{1}{c|}{98.1} & \multicolumn{1}{c|}{80.6} & \multicolumn{1}{c|}{90.1} & \multicolumn{1}{c|}{99.0} & \multicolumn{1}{c|}{99.2} & 92.1 \\ \hline
\multicolumn{1}{|c|}{\multirow{3}{*}{\begin{tabular}[c]{@{}c@{}}Finetune with different \\ inital learning rate\end{tabular}}} &  $0.02$ & \multicolumn{1}{c|}{99.5} & \multicolumn{1}{c|}{98.4} & \multicolumn{1}{c|}{\textbf{99.4}} & \multicolumn{1}{c|}{98.3} & \multicolumn{1}{c|}{85.6} & \multicolumn{1}{c|}{\textbf{93.8}} & \multicolumn{1}{c|}{99.3} & \multicolumn{1}{c|}{99.1} & \textbf{95.2} \\ [3pt] \cline{2-2}
\multicolumn{1}{|c|}{} &  $0.01$ & \multicolumn{1}{c|}{\textbf{99.6}} & \multicolumn{1}{c|}{\textbf{98.7}} & \multicolumn{1}{c|}{99.3} & \multicolumn{1}{c|}{\textbf{98.3}} & \multicolumn{1}{c|}{\textbf{86.8}} & \multicolumn{1}{c|}{93.5} & \multicolumn{1}{c|}{\textbf{99.4}} & \multicolumn{1}{c|}{\textbf{99.3}} & 94.8 \\ [3pt] \cline{2-2}
\multicolumn{1}{|c|}{} &  $0.005$ & \multicolumn{1}{c|}{99.3} & \multicolumn{1}{c|}{98.4} & \multicolumn{1}{c|}{99.1} & \multicolumn{1}{c|}{98.1} & \multicolumn{1}{c|}{86.5} & \multicolumn{1}{c|}{93.6} & \multicolumn{1}{c|}{99.1} & \multicolumn{1}{c|}{99.2} & 95.0 \\ [3pt] \hline
\end{tabular}
\label{table_5}
\end{table*}

\begin{table*}[]
\centering
\caption{Results of fine-tuning our traffic light detection model that are pre-trained using different amounts of generated data.}
\begin{tabular}{|cc|ccccccccc|}
\hline
\multicolumn{2}{|c|}{\multirow{3}{*}{Method}} & \multicolumn{9}{c|}{$AP_{50}$} \\ \cline{3-11} 
\multicolumn{2}{|c|}{} & \multicolumn{3}{c|}{Color} & \multicolumn{2}{c|}{Validity} & \multicolumn{2}{c|}{Temporal state} & \multicolumn{2}{c|}{Light state} \\ \cline{3-11} 
\multicolumn{2}{|c|}{} & \multicolumn{1}{c|}{\texttt{red}} & \multicolumn{1}{c|}{\texttt{yellow}*} & \multicolumn{1}{c|}{\texttt{green}} & \multicolumn{1}{c|}{\texttt{active}} & \multicolumn{1}{c|}{\texttt{inactive}*} & \multicolumn{1}{c|}{\texttt{flashing}*} & \multicolumn{1}{c|}{\texttt{non-flashing}} & \multicolumn{1}{c|}{\texttt{on}} & \texttt{off}* \\ \hline
\multicolumn{2}{|c|}{Baseline} & \multicolumn{1}{c|}{99.3} & \multicolumn{1}{c|}{98.0} & \multicolumn{1}{c|}{98.9} & \multicolumn{1}{c|}{98.1} & \multicolumn{1}{c|}{80.6} & \multicolumn{1}{c|}{90.1} & \multicolumn{1}{c|}{99.0} & \multicolumn{1}{c|}{99.2} & 92.1 \\ \hline
\multicolumn{1}{|c|}{\multirow{10}{*}{\begin{tabular}[c]{@{}c@{}}Pretrain with \\ different amounts \\ of synthetic data\end{tabular}}} & 100\% & \multicolumn{1}{c|}{99.6} & \multicolumn{1}{c|}{98.7} & \multicolumn{1}{c|}{99.3} & \multicolumn{1}{c|}{98.3} & \multicolumn{1}{c|}{86.8} & \multicolumn{1}{c|}{93.5} & \multicolumn{1}{c|}{99.4} & \multicolumn{1}{c|}{99.3} & 94.8 \\ \cline{2-2}
\multicolumn{1}{|c|}{} & 90\% & \multicolumn{1}{c|}{99.5} & \multicolumn{1}{c|}{98.9} & \multicolumn{1}{c|}{99.5} & \multicolumn{1}{c|}{98.5} & \multicolumn{1}{c|}{87.0} & \multicolumn{1}{c|}{\textbf{94.0}} & \multicolumn{1}{c|}{\textbf{99.6}} & \multicolumn{1}{c|}{\textbf{99.6}} & 95.2 \\ \cline{2-2}
\multicolumn{1}{|c|}{} & 80\% & \multicolumn{1}{c|}{\textbf{99.6}} & \multicolumn{1}{c|}{\textbf{99.2}} & \multicolumn{1}{c|}{\textbf{99.6}} & \multicolumn{1}{c|}{\textbf{98.5}} & \multicolumn{1}{c|}{\textbf{87.4}} & \multicolumn{1}{c|}{93.8} & \multicolumn{1}{c|}{99.4} & \multicolumn{1}{c|}{99.5} & \textbf{95.4} \\ \cline{2-2}
\multicolumn{1}{|c|}{} & 70\% & \multicolumn{1}{c|}{99.6} & \multicolumn{1}{c|}{99.0} & \multicolumn{1}{c|}{99.4} & \multicolumn{1}{c|}{98.3} & \multicolumn{1}{c|}{86.9} & \multicolumn{1}{c|}{93.5} & \multicolumn{1}{c|}{99.6} & \multicolumn{1}{c|}{99.4} & 94.9 \\ \cline{2-2}
\multicolumn{1}{|c|}{} & 60\% & \multicolumn{1}{c|}{99.4} & \multicolumn{1}{c|}{98.6} & \multicolumn{1}{c|}{99.5} & \multicolumn{1}{c|}{98.3} & \multicolumn{1}{c|}{85.5} & \multicolumn{1}{c|}{93.0} & \multicolumn{1}{c|}{99.2} & \multicolumn{1}{c|}{99.3} & 94.1 \\ \cline{2-2}
\multicolumn{1}{|c|}{} & 50\% & \multicolumn{1}{c|}{99.7} & \multicolumn{1}{c|}{98.8} & \multicolumn{1}{c|}{99.3} & \multicolumn{1}{c|}{98.4} & \multicolumn{1}{c|}{85.0} & \multicolumn{1}{c|}{92.1} & \multicolumn{1}{c|}{99.0} & \multicolumn{1}{c|}{99.5} & 95.2 \\ \cline{2-2}
\multicolumn{1}{|c|}{} & 40\% & \multicolumn{1}{c|}{99.5} & \multicolumn{1}{c|}{98.4} & \multicolumn{1}{c|}{99.4} & \multicolumn{1}{c|}{98.2} & \multicolumn{1}{c|}{83.2} & \multicolumn{1}{c|}{92.5} & \multicolumn{1}{c|}{98.8} & \multicolumn{1}{c|}{99.1} & 93.8 \\ \cline{2-2}
\multicolumn{1}{|c|}{} & 30\% & \multicolumn{1}{c|}{99.3} & \multicolumn{1}{c|}{98.2} & \multicolumn{1}{c|}{99.0} & \multicolumn{1}{c|}{98.0} & \multicolumn{1}{c|}{81.9} & \multicolumn{1}{c|}{91.8} & \multicolumn{1}{c|}{99.1} & \multicolumn{1}{c|}{99.3} & 93.1 \\ \cline{2-2}
\multicolumn{1}{|c|}{} & 20\% & \multicolumn{1}{c|}{99.4} & \multicolumn{1}{c|}{98.5} & \multicolumn{1}{c|}{99.2} & \multicolumn{1}{c|}{98.3} & \multicolumn{1}{c|}{81.5} & \multicolumn{1}{c|}{90.3} & \multicolumn{1}{c|}{99.1} & \multicolumn{1}{c|}{99.3} & 92.3 \\ \cline{2-2}
\multicolumn{1}{|c|}{} & 10\% & \multicolumn{1}{c|}{99.4} & \multicolumn{1}{c|}{98.3} & \multicolumn{1}{c|}{99.1} & \multicolumn{1}{c|}{98.0} & \multicolumn{1}{c|}{80.9} & \multicolumn{1}{c|}{90.5} & \multicolumn{1}{c|}{98.9} & \multicolumn{1}{c|}{99.0} & 92.4 \\ \hline
\end{tabular}
\label{table_6}
\end{table*}
 
\subsection{Comparison with State-of-the-Art Methods}
Finally, we compare our approach with other methods that are specifically designed for tackling the data imbalance problem. We divide these methods into four types, including loss adjusting, data sampling, joint of them, and the generative methods. We report the performances of our TLR model using these methods in Table~\ref{table_7}. Our approach achieves the best results on all of the rare classes compared with the competing algorithms, indicating the advantage of synthesized data generated by TL-GAN. 
For the loss adjusting methods, the best performing model applying class-balance loss~\cite{class-Balanced_sampling} exceeds the baseline by 0.1\%(\texttt{yellow}), 0.5\%(\texttt{inactive}), 1.9\%(\texttt{flashing}) and 2.4\%(\texttt{off}). For the data sampling methods, the best performing model applying LWS~\cite{kang_Decoupling} outperforms the baseline by 0.3\%(\texttt{yellow}), 2.5\%(\texttt{inactive}), 2.2\%(\texttt{flashing}) and 2.5\%(\texttt{off}). By joint loss adjusting and data sampling, a better model, i.e. Range loss \& LWS, is obtained, outperforming the baseline by 0.5\%(\texttt{yellow}), 5.2\%(\texttt{inactive}), 3.1\%(\texttt{flashing}) and 2.5\%(\texttt{off}). For the generative methods, our method performs the best, exceeding the baseline by 1.2\%(\texttt{yellow}), 6.8\%(\texttt{inactive}), 3.7\%(\texttt{flashing}) and 3.3\%(\texttt{off}). Moreover, our approach outperforms the best competitor (Range loss \& LWS) by 0.7\%(\texttt{yellow}), 1.6\%(\texttt{inactive}), 0.6\%(\texttt{flashing}) and 0.8\%(\texttt{off}), achieving the state-of-the-art performance. Moreover, our approach is orthogonal to the loss adjusting and data sampling methods, and our performance can be further improved if having them combined.

\begin{table*}[]
\centering
\caption{Comparison with the state-of-the-art methods that are designed for tackling data imbalance.}
\scalebox{0.9}{
\begin{tabular}{llccccccccc}
\hline
\multicolumn{2}{c|}{\multirow{3}{*}{Method}} & \multicolumn{9}{c}{$AP_{50}$} \\ \cline{3-11} 
\multicolumn{2}{c|}{} & \multicolumn{3}{c|}{Color} & \multicolumn{2}{c|}{Validity} & \multicolumn{2}{c|}{Temporal state} & \multicolumn{2}{c}{Light state} \\ \cline{3-11} 
\multicolumn{2}{c|}{} & \multicolumn{1}{c|}{\texttt{red}} & \multicolumn{1}{c|}{\texttt{yellow}*} & \multicolumn{1}{c|}{\texttt{green}} & \multicolumn{1}{c|}{\texttt{active}} & \multicolumn{1}{c|}{\texttt{inactive}*} & \multicolumn{1}{c|}{\texttt{flashing}*} & \multicolumn{1}{c|}{\texttt{non-flashing}} & \multicolumn{1}{c|}{\texttt{on}} & \multicolumn{1}{c}{\texttt{off}*} \\ \hline
\multicolumn{2}{c|}{Baseline} & 99.3 & 98.0 & \multicolumn{1}{c|}{98.9} & 98.1 & \multicolumn{1}{c|}{80.6} & 90.1 & \multicolumn{1}{c|}{99.0} & 99.2 & \multicolumn{1}{c}{92.1} \\ \hline
\multicolumn{1}{l|}{\multirow{3}{*}{Loss adjusting}} & \multicolumn{1}{l|}{Focal loss~\cite{focal_loss}} & 98.9 & 97.8 & \multicolumn{1}{c|}{98.6} & 98.1 & \multicolumn{1}{c|}{81.0} & 91.6 & \multicolumn{1}{c|}{99.2} & 99.4 & \multicolumn{1}{c}{93.8} \\
\multicolumn{1}{l|}{} & \multicolumn{1}{l|}{Range loss~\cite{range_loss}} & 99.4 & 98.6 & \multicolumn{1}{c|}{99.3} & 98.0 & \multicolumn{1}{c|}{80.7} & 90.3 & \multicolumn{1}{c|}{98.8} & 99.3 & \multicolumn{1}{c}{93.7} \\
\multicolumn{1}{l|}{} & \multicolumn{1}{l|}{Class-balance loss~\cite{classbalance_loss}} & 99.3 & 98.1 & \multicolumn{1}{c|}{99.1} & 98.2 & \multicolumn{1}{c|}{81.1} & 92.0 & \multicolumn{1}{c|}{99.3} & 99.5 & \multicolumn{1}{c}{94.5} \\ \hline
\multicolumn{1}{l|}{\multirow{5}{*}{Data sampling}} & \multicolumn{1}{l|}{Over-sampling~\cite{oversample}} & 99.1 & 98.6 & \multicolumn{1}{c|}{99.4} & 97.8 & \multicolumn{1}{c|}{80.7} & 90.3 & \multicolumn{1}{c|}{98.9} & 99.3 & \multicolumn{1}{c}{93.7} \\
\multicolumn{1}{l|}{} & \multicolumn{1}{l|}{Under-sampling~\cite{undersample}} & 98.6 & 97.5 & \multicolumn{1}{c|}{98.2} & 96.9 & \multicolumn{1}{c|}{79.3} & 90.6 & \multicolumn{1}{c|}{98.5} & 99.5 & \multicolumn{1}{c}{94.6} \\
\multicolumn{1}{l|}{} & \multicolumn{1}{l|}{cRT~\cite{kang_Decoupling}} & 99.6 & 98.7 & \multicolumn{1}{c|}{99.5} & 97.8 & \multicolumn{1}{c|}{80.2} & 90.1 & \multicolumn{1}{c|}{99.1} & 98.2 & \multicolumn{1}{c}{88.6} \\
\multicolumn{1}{l|}{} & \multicolumn{1}{l|}{$\tau$-normalized~\cite{kang_Decoupling}} & 99.3 & 98.4 & \multicolumn{1}{c|}{99.3} & 98.4 & \multicolumn{1}{c|}{84.7} & 92.1 & \multicolumn{1}{c|}{99.3} & 99.5 & \multicolumn{1}{c}{94.6} \\
\multicolumn{1}{l|}{} & \multicolumn{1}{l|}{LWS~\cite{kang_Decoupling}} & 99.4 & 98.3 & \multicolumn{1}{c|}{99.4} & 98.3 & \multicolumn{1}{c|}{83.1} & 92.3 & \multicolumn{1}{c|}{99.4} & 99.5 & \multicolumn{1}{c}{94.7} \\ \hline
\multicolumn{1}{l|}{\multirow{4}{*}{Joint}} & \multicolumn{1}{l|}{Range loss \& $\tau$-normalized \textcircled{1}} & 99.5 & 98.2 & \multicolumn{1}{c|}{99.5} & 98.4 & \multicolumn{1}{c|}{84.2} & 92.3 & \multicolumn{1}{c|}{99.1} & 99.4 & \multicolumn{1}{c}{94.8} \\
\multicolumn{1}{l|}{} & \multicolumn{1}{l|}{Class-balance loss \& $\tau$-normalized \textcircled{2}} & 99.3 & 98.3 & \multicolumn{1}{c|}{99.3} & 98.3 & \multicolumn{1}{c|}{84.9} & 93.3 & \multicolumn{1}{c|}{99.2} & 99.5 & \multicolumn{1}{c}{95.0} \\
\multicolumn{1}{l|}{} & \multicolumn{1}{l|}{Range loss \& LWS \textcircled{3}} & \textbf{99.7} & 98.5 & \multicolumn{1}{c|}{99.4} & 98.5 & \multicolumn{1}{c|}{85.8} & 93.2 & \multicolumn{1}{c|}{99.3} & 99.3 & \multicolumn{1}{c}{94.6} \\
\multicolumn{1}{l|}{} & \multicolumn{1}{l|}{Class-balance loss \& LWS \textcircled{4}} & 99.5 & 98.3 & \multicolumn{1}{c|}{99.2} & 98.4 & \multicolumn{1}{c|}{82.8} & 91.8 & \multicolumn{1}{c|}{99.1} & 99.4 & \multicolumn{1}{c}{94.2} \\ \hline
\multicolumn{1}{l|}{\multirow{3}{*}{Generative method}} & \multicolumn{1}{l|}{PA-GAN~\cite{PAGAN}} & 99.0 & 97.8 & \multicolumn{1}{c|}{98.5} & 97.8 & \multicolumn{1}{c|}{80.5} & 90.4 & \multicolumn{1}{c|}{98.8} & 98.9 & \multicolumn{1}{c}{92.1} \\
\multicolumn{1}{l|}{} & \multicolumn{1}{l|}{FQ-GAN~\cite{FQ-GAN}} & 99.3 & 98.4 & \multicolumn{1}{c|}{99.3} & 98.2 & \multicolumn{1}{c|}{81.0} & 90.2 & \multicolumn{1}{c|}{99.3} & 99.1 & \multicolumn{1}{c}{93.4} \\
\multicolumn{1}{l|}{} & \multicolumn{1}{l|}{Ours} & 99.6 & \textbf{99.2} & \multicolumn{1}{c|}{\textbf{{\ul 99.6}}} & \textbf{98.5} & \multicolumn{1}{c|}{{\ul \textbf{87.4}}} & \textbf{93.8} & \multicolumn{1}{c|}{\textbf{99.4}} & \textbf{99.5} & \multicolumn{1}{c}{\textbf{95.4}} \\ \hline
 &  &  &  &  &  &  &  &  &  &  \\ \hline
\multicolumn{1}{l|}{\multirow{4}{*}{\begin{tabular}[c]{@{}c@{}}Joint ours with \\ other methods\end{tabular}}} & \multicolumn{1}{l|}{\textcircled{1} \& Ours} & 99.4 & 99.3 & \multicolumn{1}{c|}{99.3} & 98.4 & \multicolumn{1}{c|}{86.1} & 93.5 & \multicolumn{1}{c|}{99.2} & 99.5 & \multicolumn{1}{c}{95.7} \\
\multicolumn{1}{l|}{} & \multicolumn{1}{l|}{\textcircled{2} \& Ours} & 99.3 & 99.2 & \multicolumn{1}{c|}{99.6} & 98.2 & \multicolumn{1}{c|}{86.0} & {\ul 94.5} & \multicolumn{1}{c|}{{\ul 99.7}} & {\ul 99.8} & \multicolumn{1}{c}{{\ul 95.9}} \\
\multicolumn{1}{l|}{} & \multicolumn{1}{l|}{\textcircled{3} \& Ours} & {\ul 99.7} & {\ul 99.3} & \multicolumn{1}{c|}{99.2} & 98.5 & \multicolumn{1}{c|}{86.2} & 93.9 & \multicolumn{1}{c|}{99.5} & 99.4 & \multicolumn{1}{c}{95.5} \\
\multicolumn{1}{l|}{} & \multicolumn{1}{l|}{\textcircled{4} \& Ours} & 99.2 & 98.8 & \multicolumn{1}{c|}{99.5} & {\ul 98.7} & \multicolumn{1}{c|}{86.6} & 93.9 & \multicolumn{1}{c|}{99.3} & 99.2 & \multicolumn{1}{c}{95.4} \\ \hline
\end{tabular}}
\label{table_7}
\end{table*}
\section{Conclusion}
In this paper, we have proposed a generative approach TL-GAN to generate traffic light sequences for the rare classes to improve TLR for autonomous driving. TL-GAN makes the traffic light color of generated images controllable and produces the bounding box information of bulb and lightbox. Thus, the synthesized data can be used for both classification and detection tasks. 
Extensive experiments demonstrate that our approach can not only generate realistic and diverse traffic light sequences but also brings substantial improvements to the performance of TLR, in particular on the rare classes.  
While this work demonstrates the efficacy of applying TL-GAN to synthesize various types of traffic light sequences, the future work can extend this approach to generate traffic light data with more signals (e.g., digits and arrow shape) or different whether conditions (e.g., rain and snow).
\balance
\bibliographystyle{IEEEtran}
\bibliography{new_ref}
\vfill
\end{document}